\documentclass[sigconf,nonacm]{acmart}
\AtBeginDocument{%
  }

\usepackage{multirow}
\setcopyright{none}
\begin{document}

\title{SA-RSQ: A Versatile Sparse Representation Framework for Multi-modal Recommender Systems}

\author{Xiang Wang}
\affiliation{
  \institution{Tianjin University}
  \city{Tianjin}
  \country{China}
}
\email{wangx0502@tju.edu.cn}

\author{Shigang Quan}
\affiliation{
  \institution{Meituan}
  \city{Beijing}
  \country{China}
}
\email{quanshigang@meituan.com}

\author{Tingzhen Chang}
\affiliation{
  \institution{Meituan}
  \city{Beijing}
  \country{China}
}
\email{changtingzhen@meituan.com}

\author{Kang Yang}
\affiliation{
  \institution{Meituan}
  \city{Beijing}
  \country{China}
}
\email{yangkang29@meituan.com}

\author{Sitong Chen}
\authornote{Corresponding author}
\affiliation{
  \institution{Meituan}
  \city{Beijing}
  \country{China}
}
\email{chensitong10@meituan.com}

\author{Yabo Fan}
\affiliation{
  \institution{Meituan}
  \city{Beijing}
  \country{China}
}
\email{fanyabo@meituan.com}

\author{Xingxing Wang}
\affiliation{
  \institution{Meituan}
  \city{Beijing}
  \country{China}
}
\email{wangxingxing04@meituan.com}

\author{Zhaodian He}
\affiliation{
  \institution{Institute of Software Chinese Academy of Sciences}
  \city{Beijing}
  \country{China}
}
\email{hezhaodian20@mails.ucas.ac.cn}

\renewcommand{\shortauthors}{Xiang Wang et al.}


\begin{abstract}
Deploying high-dimensional multimodal features in industrial recommender systems incurs substantial storage and latency overhead. Hard quantization is compact but introduces boundary distortion, whereas dense soft quantization couples representation quality to the limited storage budget. We propose \textit{Sparse Activation-based Residual Soft Quantization (SA-RSQ)}, which uses Top-$K$ sparse routing and softmax weights to store compact \textit{(Index, Probability)} tuples. The stored tuples decouple per-item storage from codebook dimensionality; for a fixed selected support, gradients propagate through the routing weights and weighted reconstruction without relying on a straight-through estimator. Experiments on a proprietary food-delivery advertising dataset show favorable reconstruction-performance and CTR trade-offs across storage budgets of 8--48 bytes per item. A preliminary Next-Distribution Prediction study and a one-week online A/B test further demonstrate the practical potential of SA-RSQ, with relative lifts of +2.51\% in CTR and +3.66\% in CPM.
\end{abstract}




\maketitle

\section{Introduction}
Modern recommender systems are undergoing a profound paradigm shift, increasingly leveraging high-dimensional semantic features (e.g., multimodal representations of 2048 dimensions or higher) extracted by Multimodal Large Language Models (MLLMs) to enhance semantic representation capabilities, particularly for cold-start scenarios~\cite{rajput2023tiger,zheng2024adapting}. However, directly deploying these dense vectors in industrial-scale systems serving hundreds of millions of items incurs prohibitive storage overhead and inference latency bottlenecks. To alleviate this storage bottleneck, existing approaches widely adopt RQ-VAE-based cascaded hard quantization techniques \cite{jegou2010pq, oord2018vqvae, lee2022rqvae, hou2023opq, luo2024rqkmeans, wan2026r3vae} to drastically compress these features into ultra-short discrete Semantic IDs (typically requiring only 8 bytes).

Nevertheless, such extreme compression comes at the cost of irreversible semantic distortion, which is particularly detrimental to recommender systems that heavily rely on fine-grained similarity discrimination. In essence, the current landscape presents a missing middle ground between extreme compression (with severe information loss) and high-dimensional embeddings (with prohibitive storage overhead), with no principled mechanism to smoothly trade off between the two.

Among these limitations, one fundamental issue is \textit{boundary distortion}: the $\arg\min$ operation forces semantically similar items (e.g., ``iPhone 15'' and ``iPhone 15 Pro'') to either share the same discrete ID or receive entirely different ones depending on boundary proximity, erasing the continuous distance structure essential for fine-grained recommendation systems. Furthermore, increasing residual stages to compensate yields diminishing returns, as additional stages introduce more quantization noise than useful signal, ultimately causing semantic collapse~\cite{lee2022rqvae}. Beyond representation quality, hard assignment requires the Straight-Through Estimator (STE) for gradient approximation~\cite{bengio2013ste,lee2022rqvae}, whose inherent bias both destabilizes training and prevents true end-to-end alignment between representation learning and downstream objectives~\cite{takida2022sqvae}. As a consequence, the resulting Semantic IDs are frozen after generation: the quantization model optimizes a reconstruction objective while the downstream recommender pursues CTR/CVR targets, and this fundamental objective mismatch renders hard-coded SIDs perpetually misaligned with task-specific requirements. Finally, distances between discrete IDs in the symbolic space are fundamentally unmeasurable, precluding meaningful semantic similarity computation in the quantized domain.

To overcome the non-differentiability of hard quantization, recent works explore soft quantization (e.g., SoftVQ-VAE~\cite{chen2025softvqvae}), which bypasses STE by multiplying soft assignment distributions with the codebook to produce low-dimensional dense embeddings. However, under the stringent storage constraints of recommender systems (e.g., 32 bytes can store at most 16 dimensions in Float16 format), projecting 2048D MLLM representation vectors into such a narrow bottleneck imposes severe fitting pressure on the encoder, leading to severe semantic collapse; while naively increasing dimensions would proportionally inflate storage costs. This naturally raises a critical open question: \textit{Can we find an optimal Pareto trade-off between high-dimensional dense embeddings (with prohibitive storage) and discrete Semantic IDs (with irreversible distortion and non-differentiability)?}

To this end, this paper proposes \textit{\textbf{S}parse \textbf{A}ctivation-based \textbf{R}esidual \textbf{S}oft \textbf{Q}uantization (\textbf{SA-RSQ})}. Our core insight is \textbf{breaking the strong coupling between storage footprint and semantic representation ability}. We use Top-$K$ support selection followed by a masked softmax~\cite{shazeer2017topk,fedus2022switchtransformer}, yielding compact \textit{(Index, Prob)} tuples. The support selection is discrete, but for a fixed support the probabilities, codebook values, and weighted reconstruction remain differentiable; this avoids the straight-through estimator used by hard quantization. Since only sparse routing tuples are stored, the per-item footprint is decoupled from codebook dimensionality. At inference, a table lookup and probability-weighted sum reconstruct the representation.

In summary, \textit{SA-RSQ} effectively decouples storage constraints from representation dimensionality, establishing a measurable and differentiable semantic space that bridges the gap between extreme hard compression and memory-intensive dense embeddings. The core contributions of this work are summarized as follows:

\textbf{(1) Differentiable Sparse Soft Quantization:}
We propose \textit{SA-RSQ}, a sparse soft quantization framework that replaces heuristic STE-based routing with differentiable Top-$K$ sparse routing, enabling flexible storage-performance trade-offs from 8 to 48 bytes.

\textbf{(2) Superior Compression-Performance Trade-offs:}
Offline experiments demonstrate favorable trade-offs between storage efficiency and recommendation performance under the evaluated budgets. Its probabilistic representations also support a preliminary ``Next-Distribution Prediction'' formulation for generative recommendation.

\textbf{(3) Industrial Deployment Validation:}
\textit{SA-RSQ} has been deployed in a Food Delivery Advertising Platform, where online A/B tests demonstrate practical effectiveness with relative improvements of +2.51\% in CTR and +3.66\% in CPM over the production baseline.

To support reproducibility, we release the core implementation of \textit{SA-RSQ} at \url{https://github.com/WishArdently/SA-RSQ}.

\begin{figure}[t]
  \centering
  \includegraphics[width=\linewidth]{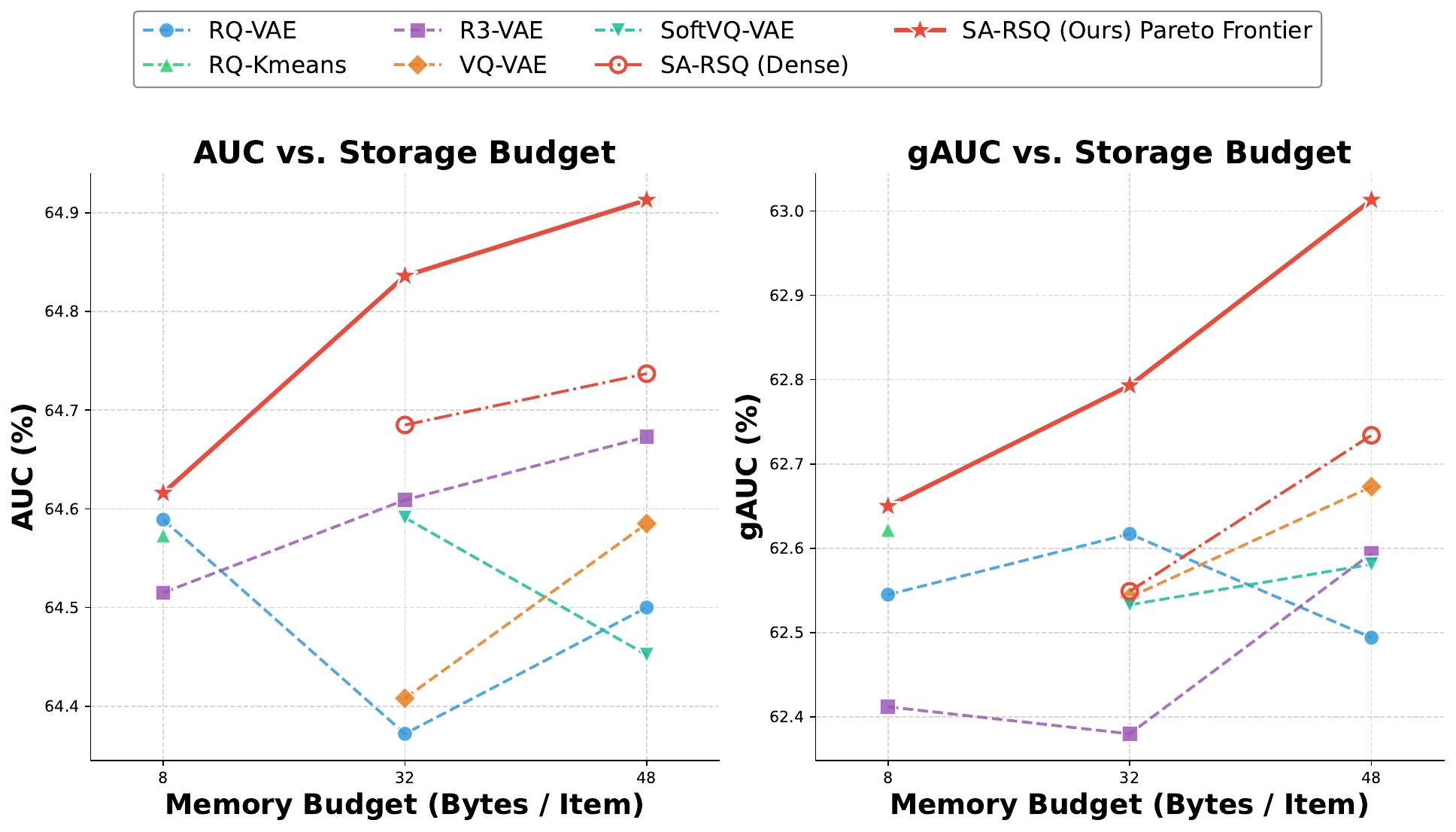}
  \caption{
  Overall performance comparison of various representation compression methods. Our \textit{SA-RSQ} achieves a favorable trade-off between compression efficiency and recommendation performance.
  }
  \label{pareto_frontier}
  \Description{A plot comparing reconstruction or recommendation performance against storage budgets for several quantization methods. SA-RSQ is shown among the evaluated configurations and follows a favorable storage-performance frontier.}
\end{figure}
\section{Related Work}
\label{sec:related_work}

\subsection{Vector Quantization for Representation Compression}

Vector quantization (VQ) has a long history in signal processing and information retrieval. Product Quantization (PQ)~\cite{jegou2010pq} decomposes vectors into disjoint sub-spaces and quantizes each independently. Optimized Product Quantization (OPQ)~\cite{Ge2014OPQ} further learns a rotation matrix to minimize distortion. In the deep learning era, VQ-VAE~\cite{oord2018vqvae} introduced a learnable discrete codebook within the VAE framework via the Straight-Through Estimator (STE)~\cite{bengio2013ste}, and RQ-VAE~\cite{lee2022rqvae} extends it to cascaded residual quantization for hierarchical semantic IDs, while R3-VAE~\cite{wan2026r3vae} learns the codebook through soft aggregation during training followed by a hard $\arg\max$ truncation to obtain semantic IDs.

In recommender systems, VQ-Rec~\cite{hou2023opq} leverages vector-quantized representations for transferable sequential recommendation. QARM \cite{luo2024rqkmeans} employs residual K-Means for industrial-scale multi-modal semantic IDs. Despite achieving extreme compression (e.g., 8 bytes per item), all these methods rely on hard assignment via $\arg\min$, which leads to (1) \textit{item collision}—semantically similar items mapped to identical codes—and (2) \textit{STE-induced optimization instability} that hinders end-to-end alignment with downstream objectives.

\subsection{Semantic IDs for Generative Recommendation}

Generative recommendation formulates item retrieval as sequence-to-sequence generation~\cite{wang2024grbase1}. TIGER~\cite{rajput2023tiger} first established the ``Next-Token Prediction'' (NTP) paradigm by autoregressively generating RQ-VAE semantic IDs. Subsequent works advance this paradigm along multiple axes: LETTER~\cite{bao2024letter} and TokenRec~\cite{qu2024tokenrec} design learnable tokenizers that integrate collaborative signals into ID construction; ETEGRec~\cite{liu2025etegrec} enables end-to-end joint optimization of tokenization and generation; OneRec~\cite{deng2025onerec} and PLUM~\cite{he2025plum} demonstrate industrial-scale deployment at Kuaishou and YouTube respectively; LongSID~\cite{hou2025longsid} and FORGE~\cite{fu2025forge} address efficiency and robustness of ID generation; OneRec-Think~\cite{liu2025onerecthink} further integrates explicit chain-of-thought reasoning into generative recommendation; and GRID~\cite{ju2025grwithsid} provides a modular framework for systematic evaluation.

Despite recent progress, the NTP paradigm inherits fundamental limitations of hard quantization: the discrete IDs create a non-differentiable boundary preventing true end-to-end optimization, and autoregressive decoding suffers from accumulation of prediction errors. These observations motivate our ``Next-Distribution Prediction'' paradigm, where the generator predicts continuous sparse probability distributions over the codebook, enabling differentiable training and graceful error tolerance.

\begin{figure*}[h]
  \centering
  \includegraphics[width=\linewidth]{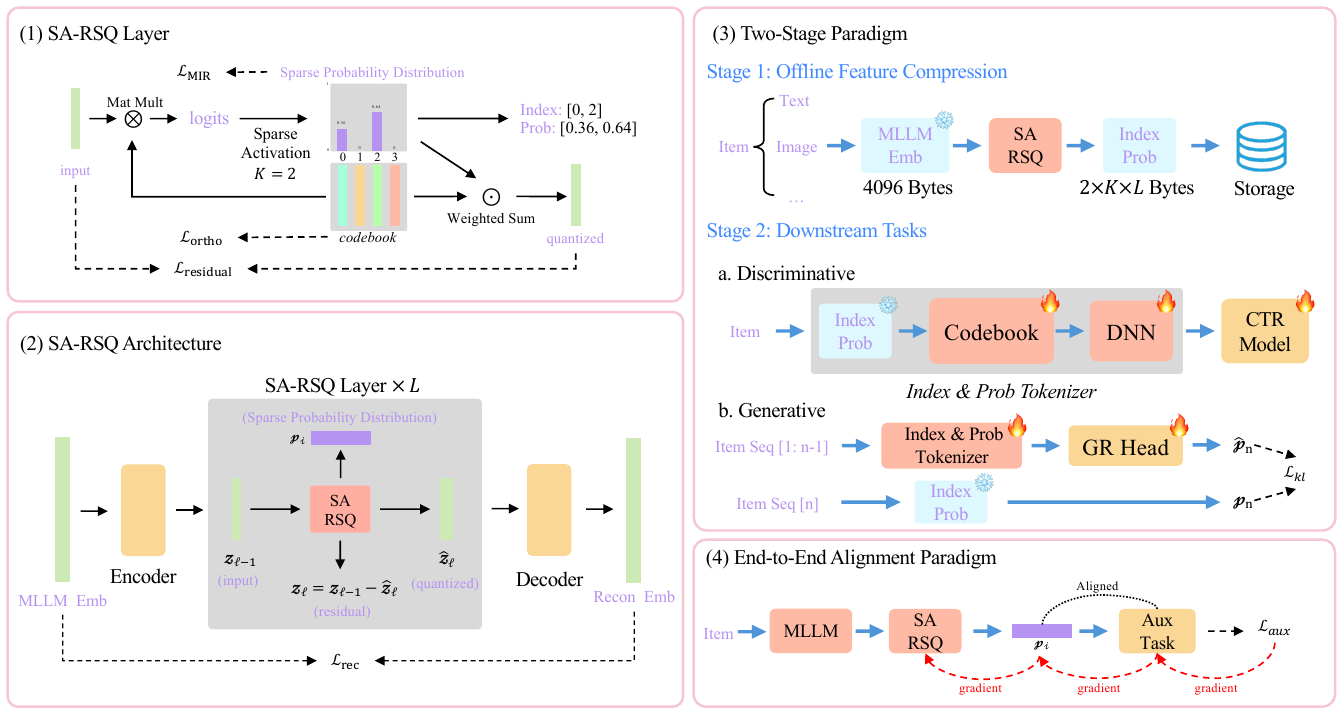}
  \caption{The overall architecture of \textit{SA-RSQ}. \textbf{(1)} The core layer utilizes differentiable Top-$k$ activation to extract sparse \textit{(Index, Prob)} tuples, reconstructing features via probability-weighted summation. \textbf{(2)} The multi-layer residual architecture for progressive quantization. \textbf{(3)} The two-stage deployment paradigm, which decouples storage from dimensionality by freezing sparse tuples while fine-tuning the codebook. \textbf{(4)} The end-to-end alignment paradigm, where sparse probabilities serve as a differentiable routing medium to backpropagate downstream gradients without STE approximations.}
  \label{main_arch}
  \Description{architecture of SA-RSQ}
\end{figure*}

\section{Methodology}
\label{sec:methodology}

In this section, we detail the \textbf{Sparse Activation-based Residual Soft Quantization (\textit{SA-RSQ})} framework. As illustrated in Figure \ref{main_arch}, \textit{SA-RSQ} compresses high-dimensional semantic embeddings while decoupling storage constraints from representation dimensionality. Its weighted reconstruction is differentiable once the Top-$k$ support is fixed.

In the following, we first formalize the core residual soft quantization paradigm and its optimization objectives. Subsequently, we elaborate on its flexible integration into downstream recommendation systems, covering both industrial two-stage deployment and end-to-end alignment.

\subsection{Differentiable Sparse Routing}
\label{sec:sa_rsq}

\vspace{1mm}
\noindent\textbf{Iterative Residual Formulation.}
\label{sec:rsq_paradigm}
Similar to standard RQ-VAE~\cite{lee2022rqvae}, our framework employs an Encoder-Decoder architecture. The input item embedding $\mathbf{x} \in \mathbb{R}^D$ is first mapped into a lower-dimensional latent space $\mathbf{h} = \text{Encoder}(\mathbf{x}) \in \mathbb{R}^d$, setting the initial residual $\mathbf{r}_0 = \mathbf{h}$. The quantization cascades through $L$ stages, with the $l$-th stage maintaining a learnable codebook $\mathbf{C}^{(l)} \in \mathbb{R}^{V \times d}$. 

The critical departure from traditional RQ-VAE lies in replacing the non-differentiable $\arg\min$ hard assignment with a continuous soft approximation $\hat{\mathbf{r}}_{l-1}$ (the exact mechanism is detailed in Section \ref{sec:sparse_activation}). The residual vector is iteratively refined, and the final reconstructed representation $\hat{\mathbf{x}}$ is generated by decoding the aggregated approximations:
\begin{gather}
    \mathbf{r}_l = \mathbf{r}_{l-1} - \hat{\mathbf{r}}_{l-1} \\
    \hat{\mathbf{x}} = \text{Decoder}\left( \sum_{l=1}^L \hat{\mathbf{r}}_{l-1} \right)
\end{gather}
This formulation preserves the cascaded compression structure. The residual updates and weighted sums are differentiable; the discrete support changes induced by Top-$k$ selection are treated as fixed within each forward pass.

\vspace{1mm}
\noindent\textbf{Sparse Activation via Truncated Softmax.}
\label{sec:sparse_activation}
To overcome the information bottleneck of hard assignment (i.e., $\arg\min$) while maintaining extreme storage efficiency, we propose a sparse soft activation mechanism. For the $l$-th residual layer, given the input residual vector $\mathbf{r}_{l-1} \in \mathbb{R}^d$ and the codebook $\mathbf{C}^{(l)} \in \mathbb{R}^{V \times d}$ containing $V$ code words, we first compute the similarity logits $\mathbf{z}^{(l)} \in \mathbb{R}^V$ between the input and all code words via scaled dot-product:
\begin{equation}
    z_i^{(l)} = \frac{\mathbf{r}_{l-1} \cdot c_i^{(l)}}{\sqrt{d}}, \quad \forall i \in \{1, 2, \dots, V\}
\end{equation}

Standard soft quantization ~\cite{chen2025softvqvae, takida2022sqvae}~ applies a dense Softmax over all code words, which incurs prohibitive storage costs. Inspired by the sparsely-gated routing mechanisms in Mixture-of-Experts (MoE)~\cite{shazeer2017topk} and sparse attention models~\cite{peters2019sparsemodel}, we instead enforce structural sparsity by retaining only the Top-$k$ most relevant code words. Specifically, we define an index set $\mathcal{I}_k$ containing the indices of the $k$ largest values in $z$, and apply a masking operation to the logits:
\begin{equation}
    \tilde{z}^{(l)}_i = 
    \begin{cases} 
    z_i^{(l)}, & \text{if } i \in \mathcal{I}_k \\ 
    -\infty, & \text{otherwise} 
    \end{cases}
\end{equation}

Subsequently, the masked logits are passed through a Softmax function to obtain a sparse probability distribution $\mathbf{p}^{(l)} \in \mathbb{R}^V$:
\begin{equation}
    p_i^{(l)} = \frac{\exp(\tilde{z}_i^{(l)})}{\sum_{j=1}^V \exp(\tilde{z}_j^{(l)})}
\end{equation}
Due to the $-\infty$ masking, the probabilities for all non-Top-$k$ code words become exactly zero. This operation projects the dense logits onto a sparse probability simplex, ensuring that exactly $k$ positions are activated. The support selection itself is discrete, while the masked-softmax weights are differentiable with respect to the selected logits.

Finally, the quantized representation for the $l$-th layer is constructed as the convex combination of the active code words weighted by their sparse probabilities:
\begin{equation}
    \hat{\mathbf{r}}_{l-1} = \sum_{i \in \mathcal{I}_k} p_i^{(l)} c_i^{(l)}
\end{equation}

During offline processing, we only need to store the $k$ active indices and their corresponding non-zero probabilities (i.e., the \textit{Index} \& \textit{Prob} format). In the reported accounting, each index is a 16-bit unsigned integer and each probability is a Float16 value; the shared codebook and model parameters are excluded from the per-item budget. Thus, an index-only tuple costs $2LK$ bytes and an \textit{Index+Prob} tuple costs $4LK$ bytes: $L{=}4,K{=}1$ gives 8 bytes, $L{=}2,K{=}4$ gives 32 bytes, and $L{=}4,K{=}3$ gives 48 bytes. This accounting assumes $V\leq 2^{16}$ and no additional per-item padding.

\vspace{1mm}
\noindent\textbf{Curriculum Sparsity Annealing.} Applying Top-$k$ masking from the outset restricts gradient flow and hinders global semantic exploration. To prevent this premature pruning, we introduce a Cosine Sparsity Annealing schedule~\cite{loshchilov2017sgdr, soviany2022curriculumlearning}. Training initiates with fully dense activation ($k(0) = V$) for unhindered updates. At step $t$, the active budget $k(t)$ dynamically decays to the target sparsity $k_{tgt}$ via a cosine factor $\eta(t)$:
\begin{gather}
    \eta(t) = \frac{1}{2} \left( 1 + \cos\left( \frac{\pi t}{T_{ann}} \right) \right) \\
    k(t) = \max \left( k_{tgt}, \left\lfloor k_{tgt} + (V - k_{tgt}) \eta(t) \right\rfloor \right)
\end{gather}
where $T_{ann}$ is the total annealing steps. This schedule balances exploration and exploitation, reaching $k_{tgt}$ before inference so that the stored support satisfies the target budget.

\subsection{Optimization Objectives}
\label{sec:optimization}

To train \textit{SA-RSQ}, we design a joint optimization objective comprising four components: global reconstruction, layer-wise residual approximation, geometric regularization, and mutual information maximization.

\vspace{1mm}
\noindent\textbf{Global and Layer-wise Reconstruction.} 
To faithfully reconstruct the original embedding $\mathbf{x}$, we apply a global Mean Squared Error (MSE) loss $\mathcal{L}_{recon}$. Furthermore, unlike hard quantization ~\cite{oord2018vqvae, lee2022rqvae} which relies on heuristic commitment losses with stop-gradients, our differentiable architecture inherently aligns the encoder and codebook. To explicitly stabilize cascaded training and prevent representation collapse, we introduce a Layer-wise Residual Reconstruction Loss ($\mathcal{L}_{residual}$) to minimize the approximation error at each stage:
\begin{gather}
    \mathcal{L}_{recon} = \|\mathbf{x} - \hat{\mathbf{x}}\|_2^2 \\
    \mathcal{L}_{residual} = \sum_{l=1}^L \|\mathbf{r}_{l-1} - \hat{\mathbf{r}}_{l-1}\|_2^2
\end{gather}
where $\hat{\mathbf{x}}$ is the final reconstructed embedding and  $\hat{\mathbf{r}}_{l-1}$ is the soft-quantized representation at stage $l$. This explicitly encourages the sparse routing to discover the optimal convex combination of codewords that tightly bounds the incoming residual.

\vspace{1mm}
\noindent\textbf{Geometric Codebook Regularization via Orthogonality.} While orthogonal regularization is widely used to disentangle representations~\cite{li2019mind, cen2020comirec}, we repurpose it as a crucial geometric constraint for our codebooks. Hard quantization ~\cite{lee2022rqvae, oord2018vqvae, luo2024rqkmeans, jegou2010pq, hou2023opq, Ge2014OPQ}~ treats codewords as rigid centroids. In contrast, \textit{SA-RSQ } treats them as basis vectors for soft linear combinations. To maximize the spanning capacity of these combinations and eliminate spatial redundancy~\cite{bansal2018orthoregular}, we apply an Orthogonal Loss to penalize the cosine similarity between distinct codewords within each codebook $\mathbf{C}^{(l)}$:
\begin{equation}
    \mathcal{L}_{ortho} = \sum_{l=1}^L \frac{1}{V(V-1)} \sum_{i \neq j} \frac{\mathbf{c}_i^{(l)} \cdot \mathbf{c}_j^{(l)}}{\|\mathbf{c}_i^{(l)}\|_2  \times \|\mathbf{c}_j^{(l)}\|_2}
\end{equation}
By enforcing mutual orthogonality, $\mathcal{L}_{ortho}$ ensures that the selected Top-$k$ codewords are highly independent, thereby maximizing the expressiveness of the reconstructed continuous space.

\vspace{1mm}
\noindent\textbf{Codebook Optimization via Information Maximization.}
\label{sec:mir_loss}
A critical challenge in quantization is codebook collapse~\cite{oord2018vqvae}, where only a small fraction of codewords are utilized. Instead of relying on heuristic, non-differentiable tricks like ``codebook restarts''~\cite{razavi2019vqvae2}, we propose a principled and differentiable solution based on Mutual Information Regularization (MIR).

Following information-theoretic paradigms~\cite{krause2010mirclustering, hu2017mirdiscrete}, we maximize the mutual information $I(\mathbf{X}; \mathbf{Z})$ between the inputs $\mathbf{X}$ and the soft assignments $\mathbf{Z}$. This equates to maximizing the marginal batch entropy $H(\mathbf{Z})$ while minimizing the conditional sample entropy $H(\mathbf{Z}|\mathbf{X})$. Formally, let $p_{i,j}$ denote the probability of the $i$-th sample activating the $j$-th codeword in a batch of size $B$. The sample entropy loss is defined as:
\begin{equation}
    \mathcal{L}_{sample} = -\frac{1}{B} \sum_{i=1}^B \sum_{j=1}^V p_{i,j} \log p_{i,j}
\end{equation}
Minimizing $\mathcal{L}_{sample}$ enforces \textit{microscopic sparsity}, encouraging sharp, highly discriminative distributions for individual items. Conversely, the batch entropy loss relies on the mean activation probability $\bar{p}_j = \frac{1}{B}\sum_{i=1}^B p_{i,j}$:
\begin{equation}
    \mathcal{L}_{batch} = -\sum_{j=1}^V \bar{p}_j \log \bar{p}_j
\end{equation}
Maximizing $\mathcal{L}_{batch}$ enforces \textit{macroscopic uniformity}, ensuring all codewords are explored evenly. The overall MIR loss balances both objectives:
\begin{equation}
    \mathcal{L}_{MIR} = \mathcal{L}_{sample} - \alpha \mathcal{L}_{batch}
\end{equation}
where $\alpha > 0$ is a hyper-parameter. 

While Top-$k$ masking blocks gradient flow for unselected logits, our Cosine Sparsity Annealing inherently resolves this: a large $k(t)$ in early training enables $\mathcal{L}_{batch}$ to uniformly distribute codewords across the latent space, ensuring that as $k(t)$ decays toward $k_{tgt}$, diverse items naturally activate distinct codewords at the batch level, sustaining effective gradient coverage throughout training.

\vspace{1mm}
\noindent\textbf{Overall Training Objective.}
Combining the aforementioned components, the total loss function for the first-stage pre-training of \textit{SA-RSQ} is formulated as:
\begin{equation}
    \mathcal{L}_{total} = \mathcal{L}_{recon} + \lambda_1 \mathcal{L}_{residual} + \lambda_2 \mathcal{L}_{ortho} + \lambda_3 \mathcal{L}_{MIR}
\end{equation}
where $\lambda_1, \lambda_2,$ and $\lambda_3$ are hyper-parameters controlling the relative importance of residual approximation, geometric independence, and information-theoretic regularization, respectively.

\subsection{Integration with Recommender Systems}
 \label{sec:integration}
 
\vspace{1mm}
\noindent\textbf{Two-Stage Paradigm.}
As illustrated in Figure \ref{main_arch} (2), to meet the stringent latency and memory constraints of industrial recommender systems~\cite{covington2016deep, zhou2018din, naumov2019dlrm}, we deploy \textit{SA-RSQ} via a two-stage paradigm. In the offline phase, we extract and freeze the sparse routing tuples $(Index, Prob)$ for all items. Remarkably, this guarantees a strict storage upper bound of $\mathcal{O}(k \times L)$ bytes per item. During the online phase, the downstream model merely performs lightweight lookups and weighted summations over a trainable codebook. Consequently, this paradigm enables CTR models to leverage high-fidelity 2048D semantics while maintaining a negligible memory footprint.

\vspace{1mm}
\noindent\textbf{End-to-End Alignment Paradigm.}
A fundamental flaw of traditional discrete IDs ~\cite{lee2022rqvae,oord2018vqvae,luo2024rqkmeans}~ is the optimization misalignment: the quantization model is optimized for reconstruction, which is agnostic to downstream recommendation objectives. As depicted in Figure \ref{main_arch} (4), \textit{SA-RSQ} addresses this via the End-to-End Alignment Paradigm. The sparse probabilities $p_{d,i}$ serve as a differentiable routing medium. For a fixed Top-$k$ support, downstream gradients backpropagate through the weighted sum and update the upstream encoder and codebook without a straight-through estimator~\cite{bengio2013ste}; gradients do not differentiate through changes in the discrete support.

In practice, we introduce an offline \textit{Proxy Co-training} module as a lightweight alignment instantiation. Following collaborative alignment paradigms ~\cite{xie2022eider, wu2021sgl, qiu2022contrastive}, we apply an auxiliary contrastive loss on item pairs. Because the routing weights are differentiable for a fixed support, this proxy signal fine-tunes the sparse probabilities and injects task-specific collaborative signals before downstream deployment (Section \ref{sec:ablation}).

\begin{table*}[ht]
\caption{Overall performance comparison of various representation compression methods on downstream CTR prediction. We strictly align the storage budget (\textit{Bytes/Item}) to evaluate the optimal trade-off between memory efficiency and recommendation accuracy. The best and second-best results are highlighted in bold and \underline{underlined}, respectively. }
\label{tab:main_table}
\centering
\begin{tabular}{ccccccccc}
\toprule
\textbf{Bytes/Item} &\textbf{ Method }&\textbf{ Format Details} &\textbf{ AUC(\%)$\uparrow$} & \textbf{gAUC (\%) $\uparrow$} & \textbf{RL $\downarrow$} & \textbf{PosSC $\uparrow$} & \textbf{NegSC $\downarrow$} & \textbf{SC $\uparrow$} \\
\midrule
4096 & Qwen3-VL (Original) & 2048D Dense & OOM & OOM & - & - & - & - \\
\midrule
\multirow{4}{*}{8} 
 & RQ-VAE & 4-layer SID & \underline{64.589} & 62.545 & 0.4010 & 0.5276 & \textbf{0.0046} & 0.5230 \\
 & R-Kmeans & 4-layer SID & 64.573 & \underline{62.622} & 0.5336 & 0.6907 & 0.1831 & 0.5076 \\
 & R3-VAE & 4-layer SID & 64.515 & 62.412 & \underline{0.3279} & \underline{0.7805} & \underline{0.0605}  & \textbf{0.7200 }\\
 & \textbf{SA-RSQ (Ours)} & \textbf{L=4, K=1 (Index)} & \textbf{64.616} & \textbf{62.650} & \textbf{0.2394} & \textbf{0.8535} & 0.1541 & \underline{0.6994}\\
\midrule
\multirow{6}{*}{32} 
 & RQ-VAE & 16D Dense Emb & 64.372 & \underline{62.617} & 0.3934 & 0.8751 & 0.0100 & 0.8651 \\
 & R3-VAE & 16D Dense Emb & 64.609 & 62.380 & 0.3461 & \textbf{0.9961} & 0.8692 & 0.1269 \\
 & VQ-VAE & 16D Dense Emb & 64.408 & 62.542 & 0.6215 & 0.8343 & 0.0146 & 0.8197 \\
 & SoftVQ-VAE & 16D Dense Emb & 64.452 & 62.533 & 0.4430 & 0.9673 & 0.1592 & 0.8081 \\
 & \textbf{SA-RSQ (Ours)} & \textbf{16D Dense Emb }& \underline{64.685} & 62.549 & \underline{0.3119}  & 0.8944 & \underline{0.0046} & \underline{0.8898} \\
 & \textbf{SA-RSQ (Ours)} & \textbf{L=2, K=4 (Index+Prob)} & \textbf{64.836} & \textbf{62.793} & \textbf{0.1660} & \underline{0.9055}  & \textbf{0.0012} &\textbf{ 0.9043} \\
\midrule
\multirow{6}{*}{48} 
 & RQ-VAE & 24D Dense Emb & 64.500 & 62.494 & 0.3962 & 0.8318 & 0.0094 & 0.8224 \\
 & R3-VAE & 24D Dense Emb & 64.673 & 62.594 & 0.3343 &\textbf{ 0.9902} & 0.8635 & 0.1267 \\
 & VQ-VAE & 24D Dense Emb & 64.585 & 62.673 & 0.6301 & 0.8147 & 0.0119 & 0.8028 \\
& SoftVQ-VAE & 24D Dense Emb & 64.591 & 62.581 & 0.4334 & 0.9571 & 0.1130 & 0.8441 \\
 & \textbf{SA-RSQ (Ours)} & \textbf{24D Dense Emb} & \underline{64.737} & \underline{62.734} & \underline{0.2653} & 0.9190 & \underline{0.0061} & \underline{0.9129} \\
 & \textbf{SA-RSQ (Ours)} &\textbf{L=4, K=3 (Index+Prob) }& \textbf{64.913} & \textbf{63.013} &\textbf{ 0.1553} & \underline{0.9205}  & \textbf{0.0026 }&\textbf{ 0.9179} \\
\bottomrule
\end{tabular}
\end{table*}

\section{Experiments}
In this section, we conduct comprehensive experiments to evaluate the effectiveness of \textit{SA-RSQ}. We systematically benchmark its performance against state-of-the-art quantization methods under strictly controlled storage budgets. Furthermore, we provide in-depth ablation studies to validate its core mechanisms, explore its potential in generative recommendation paradigms, and report its commercial gains in real-world online A/B tests.
\subsection{Experiment Setup}
\label{sec:exp_setup}
\vspace{1mm}
\noindent\textbf{Dataset.}
We evaluate our framework on a large-scale industrial dataset from a food-delivery advertising platform, comprising hundreds of millions of items. The raw item semantics are 2048D vectors extracted by a pre-trained MLLM~\cite{bai2025qwen3vl}. The dataset, user/item identifiers, exact interaction counts, density, and partition cardinalities are proprietary and cannot be released; consequently, this paper reports scale and dimensionality but not those exact statistics. We do not claim that the results transfer to public benchmarks without further evaluation.

\vspace{1mm}
\noindent\textbf{Downstream Model.}
Since \textit{SA-RSQ} provides model-agnostic item representations, we adopt the Deep Interest Network (DIN)~\cite{zhou2018din} as a representative backbone. To isolate information retention from downstream capacity, we unify the final item representation dimension to $D_{model}=16$ for all compression formats.

For 32/48-byte budgets, we report the reconstructed continuous embeddings (Dense Emb) of hard-quantization baselines rather than excessively long SIDs. This is an upper-bound-style comparison under relaxed storage constraints, while the sparse tuple rows use the explicit per-item accounting in Section~\ref{sec:sa_rsq}. All methods use the same downstream backbone and final dimension.

\vspace{1mm}
\noindent\textbf{Evaluation Metrics.}
We evaluate downstream CTR prediction with \textbf{AUC} and \textbf{gAUC}. We report \textbf{Reconstruction Loss (RL)}, the MSE between original and reconstructed embeddings, and \textbf{Semantic Cohesion (SC)}~\cite{wan2026r3vae}. SC is the difference between positive-pair similarity (PosSC) and negative-pair similarity (NegSC). The tables report point estimates from the common evaluation pipeline; repeated-seed standard deviations and confidence intervals are not available in the current production evaluation.

\subsection{Main Results in Discriminative Task}
\label{sec:main_results}

Table \ref{tab:main_table} summarizes the downstream CTR performance under strictly aligned storage budgets (\textit{Bytes/Item}). \textit{SA-RSQ} consistently outperforms all baselines, yielding three key observations:

\vspace{1mm}
\noindent\textbf{Superiority under Extreme Compression (8 Bytes).}
Even when constrained to store only discrete \textit{Indices} (discarding probabilities during inference), \textit{SA-RSQ} (${L=1, K=4}$) achieves the highest AUC (64.616) among 8-byte baselines. This indicates that our sparse soft routing mechanism cultivates a fundamentally more expressive codebook than the rigid assignments or suboptimal approximations prevalent in prior quantization methods~\cite{lee2022rqvae, oord2018vqvae, luo2024rqkmeans, wan2026r3vae}

\vspace{1mm}
\noindent\textbf{The Power of Sparse Probabilities (32 \& 48 Bytes).}
When budgets allow the storage of complete \textit{(Index, Prob)} tuples, \textit{SA-RSQ} shows consistent gains in the reported configurations. At 32 bytes, \textit{SA-RSQ} ($L=2, K=4$) reaches AUC \textbf{64.836}, above the 16D Dense Embedding baselines ($\approx 64.6$). At 48 bytes, it reaches the highest reported AUC (\textbf{64.913}) and gAUC (\textbf{63.013}). These results are consistent with the hypothesis that sparse routing preserves more of the high-dimensional semantic space than the evaluated dense baselines~\cite{chen2025softvqvae, lee2022rqvae, wan2026r3vae, oord2018vqvae}.

\vspace{1mm}
\noindent\textbf{High-Fidelity Reconstruction Drives Accuracy.}
We observe a clear negative correlation between RL and CTR performance. Although traditional methods ~\cite{lee2022rqvae,wan2026r3vae,oord2018vqvae,chen2025softvqvae}~ suffer irreversible information loss (RL $> 0.3$), \textit{SA-RSQ} drastically reduces RL to \textbf{0.1553} (at 48 bytes). Powered by superior structural metrics (PosSC, NegSC), \textit{SA-RSQ}  faithfully preserves continuous fine-grained semantics. This effectively mitigates the "item collision" problem, providing downstream models with richer and more accurate collaborative filtering signals.

Across the configurations reported in Table~\ref{tab:main_table}, \textit{SA-RSQ} provides a favorable storage--accuracy trade-off. The figure and table describe a frontier over the evaluated configurations; they do not establish global Pareto optimality over all possible methods or hyperparameters.

\subsection{Exploratory Study: Generative Recommendation}
\label{sec:exp_generative}

While the primary focus of this work lies in feature compression for discriminative CTR models, we conduct a preliminary exploratory study to validate the potential of \textit{SA-RSQ} in Generative Recommender Systems (GR). 

\vspace{1mm}
\noindent\textbf{Setup.} 
We adopt a standard Transformer-based sequential architecture ~\cite{zhai2024hstu}~ and train it on a million-scale industrial user interaction dataset, which is \textbf{sampled from the aforementioned} Food Delivery Advertising Platform. We compare two generative paradigms:
1) \textbf{Next Token Prediction (NTP)}: The model autoregressively predicts discrete SIDs using standard Cross-Entropy. For fair comparison, all baselines ~\cite{wan2026r3vae, luo2024rqkmeans, lee2022rqvae}~ and our \textit{SA-RSQ} use a 4-layer hard discrete configuration ($L=4, K=1$).
2) \textbf{Next Distribution Prediction (NDP)}: the model predicts continuous distributions optimized via KL divergence against \textit{SA-RSQ}'s sparse routing tuples ($L=2, K=4$). The target Top-$k$ support is fixed when computing this loss.

\vspace{1mm}
\noindent\textbf{Preliminary Results.} 
The generative retrieval performance ( Recall@$K$ and NDCG@$K$) is reported in Table \ref{tab:ntp_nvp}. Under the traditional NTP paradigm, \textit{SA-RSQ}'s discrete indices already outperform other hard quantization baselines, indicating a superior underlying codebook quality. More interestingly, when transitioning to the proposed NDP paradigm, \textit{SA-RSQ} achieves further performance gains across all metrics (e.g., R@10 improves from 0.0088 to 0.0096). 

These preliminary results suggest that probabilistic targets are feasible in this setup. They are not evidence that NDP is a complete or generally superior generative recommendation paradigm.

\begin{table}[ht]
\centering
\caption{Preliminary results on Generative Recommendation. NTP denotes traditional Next Token Prediction (using discrete IDs), while NDP denotes our proposed Next Distribution Prediction (using sparse probabilities).}
\label{tab:ntp_nvp}
\begin{tabular}{clcccc}
\toprule
\textbf{Paradigm} & \textbf{Method} & \textbf{R@10} & \textbf{N@10} & \textbf{R@20} & \textbf{N@20} \\
\midrule
 \multirow{4}{*}{NTP}
 & RQ-VAE          & 0.0068 & 0.0045 & 0.0105 & 0.0045 \\
 & R-Kmeans        & 0.0057 & 0.0031 & 0.0086 & 0.0038 \\
 & R3-VAE          & 0.0075 & 0.0043 & 0.0105 & 0.0051 \\
 & \textbf{SA-RSQ} & 0.0088 & 0.0050 & 0.0126 & 0.0060\\ 
\midrule
NDP 
 & \textbf{SA-RSQ} & 0.0096 & 0.0059 & 0.0131 & 0.0072 \\
\bottomrule
\end{tabular}
\end{table}

\subsection{Ablation Study}
\label{sec:ablation}

To deeply understand the contribution of each component in the \textit{SA-RSQ}  framework, we conduct a comprehensive ablation study by removing six key modules respectively. The results are summarized in Table \ref{tab:unified_ablation}.

\begin{table}[h]
\caption{Comprehensive ablation study of SA-RSQ. `Usage' represents the codebook utilization rate.}
\label{tab:unified_ablation}
\centering
\resizebox{\columnwidth}{!}{
\begin{tabular}{lcccc}
\toprule
\textbf{Variant} & \textbf{AUC (\%) $\uparrow$} & \textbf{RL $\downarrow$} & \textbf{Usage (\%) $\uparrow$} & \textbf{SC $\uparrow$} \\
\midrule
w/o $\mathcal{L}_{residual}$    & 64.622 & 0.3212 & 73.2 & 0.7997 \\  
w/o $\mathcal{L}_{ortho}$       & 64.565 & 0.2720 & 68.4 & 0.8005 \\ 
w/o $\mathcal{L}_{MIR}$         & 64.587 & 0.2413 & 24.5 & 0.2446 \\
w/o Proxy Co-training           & 64.746 & 0.1419 & 99.3 & 0.1014 \\
w/o Curriculum Learning         & 64.741 & 0.4942 & 15.9 & 0.1587 \\
w/o top-$k$ Mask                & 64.604 & 0.2515 & 49.7 & 0.1035 \\
\midrule
\textbf{Full Model (SA-RSQ)}    & \textbf{64.913} & \textbf{0.1553} & \textbf{100\%} & \textbf{0.9179} \\
\bottomrule
\end{tabular}
}
\end{table}
\noindent\textbf{Objective Functions.} 
Removing the layer-wise residual loss ($\mathcal{L}_{residual}$) drops the AUC to 64.622 and increases RL to 0.3212, confirming that deep supervision is vital for stabilizing cascaded quantization. Discarding the orthogonality loss ($\mathcal{L}_{ortho}$) similarly degrades performance as codewords lose their mutually exclusive representation power. Crucially, removing the Mutual Information Regularization ($\mathcal{L}_{MIR}$) triggers a catastrophic codebook collapse, with Usage plummeting to 24.5\% and SC dropping to 0.2446. This strongly validates that MIR is indispensable for maintaining macroscopic uniformity and preventing dead codes.

\vspace{1mm}
\noindent\textbf{Alignment \& Routing Mechanisms.} 
Removing the lightweight Proxy Co-training causes a severe drop in Semantic Cohesion (SC) from 0.9179 to 0.1014, alongside an AUC decline. This result is consistent with proxy contrastive learning injecting collaborative signals into the offline compression phase. Furthermore, without Curriculum Learning, the model fails to explore the semantic space, leading to the worst RL (0.4942) and 15.9\% Usage. Finally, removing the Top-$k$ mask (degenerating to dense soft quantization) degrades AUC to 64.604. These results support the contribution of the routing and annealing components in the evaluated setting.

\subsection{Online A/B Tests}
\label{app:ab_test}

To measure deployment impact, we conducted a one-week A/B test on the food-delivery advertising platform using 10\% of production traffic. We evaluated several configurations for each method online and report the peak configuration for each method in Table~\ref{tab:ab_test}, relative to a production baseline without multimodal features. \textit{SA-RSQ} produced relative lifts of +2.51\% in CTR and +3.66\% in CPM. Traffic counts, confidence intervals, p-values, guardrail metrics, and the number of online trials are withheld by the production system; the reported peak values should therefore be interpreted as deployment evidence rather than a significance analysis. We also report storage but do not provide p99 latency or throughput measurements, which remain important deployment limitations.

\begin{table}[h]
\caption{Online A/B test results. We report the peak online improvements for each method across various configurations, measured against a production baseline without multi-modal features.}
\label{tab:ab_test}
\centering
\resizebox{\columnwidth}{!}{
\begin{tabular}{lccc}
\toprule
\textbf{Method} & \textbf{Optimal Online Config} & \textbf{CTR Imp.(\%)} &\textbf{CPM Imp.(\%)} \\
\midrule
RQ-VAE           & 4-layer SID   & 0.96 & 1.27 \\
R3-VAE           & 24D Dense Emb & 0.85 & 0.97 \\
VQ-VAE           & 24D Dense Emb & 1.26 & 1.68 \\
R-Kmeans         & 4-layer SID   & 0.91 & 1.21 \\
\midrule
\textbf{SA-RSQ}  & \textbf{$L=4, K=3$} & \textbf{2.51} & \textbf{3.66} \\
\bottomrule
\end{tabular}
}
\end{table}

\section{Conclusion}
\label{sec:conclusion}

In this paper, we propose \textbf{Sparse Activation-based Residual Soft Quantization (\textit{SA-RSQ})} for compressing high-dimensional MLLM embeddings in industrial recommender systems. Top-$k$ support selection with masked-softmax weights produces compact \textit{(Index, Prob)} tuples; for a fixed support, weighted reconstruction and routing probabilities remain differentiable without a straight-through estimator. On the proprietary dataset, SA-RSQ provides favorable trade-offs across the evaluated 8--48 byte configurations. A preliminary generative study and a one-week online A/B test suggest practical potential, while the lack of public data, repeated-run uncertainty, and detailed latency statistics limits the strength of broader claims.

Future work should evaluate NDP on public or shareable benchmarks, report uncertainty and system-level latency, and study robustness across domains before treating distribution prediction as a general generative recommendation paradigm.

\bibliographystyle{ACM-Reference-Format}
\balance
\bibliography{reference}

\end{document}